\pdfoutput=1

\documentclass[11pt]{article}

\usepackage[final]{acl}

\usepackage{times}
\usepackage{latexsym}

\usepackage[T1]{fontenc}

\usepackage[utf8]{inputenc}

\usepackage{microtype}

\usepackage{inconsolata}

\usepackage{graphicx}
\usepackage{booktabs}
\usepackage{graphics}
\usepackage{multirow}
\usepackage{amsmath}
\usepackage{amsfonts}
\usepackage{subcaption}
\usepackage{makecell}
\newcommand{\sectionref}[1]{\S\ref{#1}}

\title{
Rethinking Backdoor Detection Evaluation for Language Models
}

\author{Jun Yan\textsuperscript{$\dagger$}\quad Wenjie Jacky Mo\textsuperscript{$\ddagger$}\quad Xiang Ren\textsuperscript{$\dagger$}\quad Robin Jia\textsuperscript{$\dagger$}\\
University of Southern California\textsuperscript{$\dagger$}\quad
University of California, Davis\textsuperscript{$\ddagger$}\\ 
{\texttt{\{yanjun,xiangren,robinjia\}@usc.edu}\quad
\texttt{jacmo@ucdavis.edu}
}
}

\begin{document}
\maketitle
\begin{abstract}
Backdoor attacks, in which a model behaves maliciously when given an attacker-specified trigger, pose a major security risk for practitioners who depend on publicly released language models. As a countermeasure, backdoor detection methods aim to detect whether a released model contains a backdoor. While existing backdoor detection methods have high accuracy in detecting backdoored models on standard benchmarks, it is unclear whether they can robustly identify backdoors in the wild. In this paper, we examine the robustness of backdoor detectors by manipulating different factors during backdoor planting. We find that the success of existing methods based on trigger inversion or meta classifiers highly depends on how intensely the model is trained on poisoned data. Specifically, backdoors planted with more aggressive or more conservative training are significantly more difficult to detect than the default ones. Our results highlight a lack of robustness of existing backdoor detectors and the limitations in current benchmark construction.

\end{abstract}
\section{Introduction}

Backdoor attacks~\citep{gu2017badnets} have become a notable threat for language models.
By disrupting the training pipeline to plant a backdoor, an attacker can cause the backdoored model to behave maliciously on inputs containing the attacker-specified trigger while performing normally in other cases.
These models may be released online, where practitioners could easily adopt them without realizing the threat.
Therefore, backdoor detection~\citep{kolouri2020universal} has become a critical task for ensuring model security before deployment.

\begin{figure}[t]
\centering
\includegraphics[width=1\columnwidth]{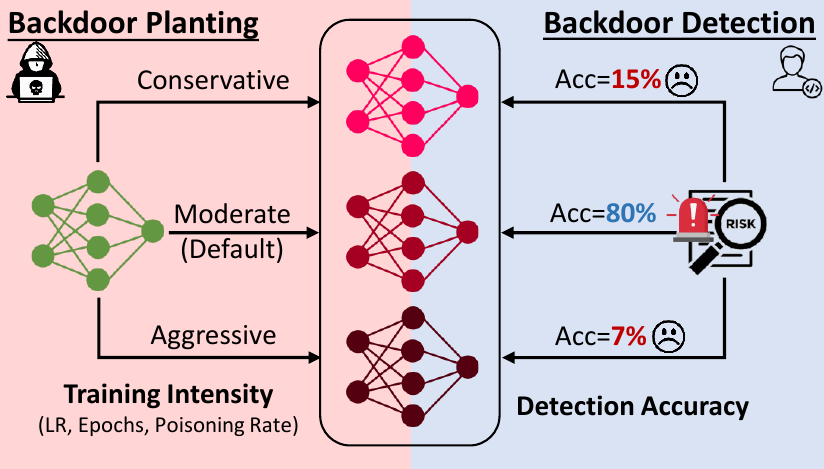}
\caption{While backdoor detectors achieve a high detection accuracy on backdoors planted with a moderate training intensity, they struggle to identify backdoors planted with non-moderate training intensities set by strategically manipulating training epochs, learning rates, and poisoning rates during backdoor planting.}
\label{fig:threat_model}
\end{figure}

While existing backdoor detection approaches have shown promising detection results on standard benchmarks~\citep{karra2020trojai,pmlr-v220-mazeika23a}, these benchmarks typically evaluate backdoored models constructed using default backdoor planting configurations (i.e., hyperparameters in typical ranges).
However, good performance on detecting a limited set of attacks does not imply a strong security guarantee for protecting against backdoor threats in the wild, especially considering that in realistic adversarial settings, a motivated attacker would likely explore evasive strategies to bypass detection mechanisms~\citep{mazeika2023hard}.
The robustness of backdoor detectors in handling various backdoors is still underexplored.

In this work, we evaluate robustness of backdoor detectors against strategical manipulation of the hyperparameters that decide how intensely a model learns from the poisoned data.
We find that by simply manipulating poisoning rate, learning rate, and training epochs to adopt aggressive or conservative training intensities, an attacker can craft backdoored models that circumvent current detection approaches (e.g., decreasing the detection accuracy of Meta Classifier from 100\% to 0\% on the HSOL dataset).
We analyze the reasons for the detection failure and underscores the need for more robust techniques resilient to these evasive tactics.

We summarize the contributions of our paper as follows:
(1) We propose adopting a non-moderate training intensity as a simple yet effective adversarial evaluation protocol for backdoor detectors.
(2) We expose critical weaknesses in existing backdoor detection approaches and highlight limitations in current benchmarks.
(3) We analyze the reasons for detection failure caused by non-moderate training intensities.
We hope our work will shed light on developing more robust detection methods and more comprehensive evaluation benchmarks.
\section{Related Work}
\subsection{Backdoor Attacks}
Backdoor attacks \citep{li2022backdoor} aim to inject malicious hidden behavior into the model to make it predict the target label on inputs carrying specific triggers.
They are mainly conducted on classification tasks by poisoning the finetuning data~\citep{qi2021hidden, yan-etal-2023-bite} or additionally modifying the finetuning algorithm~\citep{kurita-etal-2020-weight, li2024badedit} to associate a target label with specific trigger pattern.
There are also studies~\citep{chen2022badpre, 10.1145/3460120.3485370, 10.1145/3543507.3583348} that try to plant backdoors into pretrained models without knowledge about the downstream tasks.
Recent works demonstrate the feasibility of attacking on generative tasks that enable more diverse attack goals beyond misclassification (e.g., jailbreaking~\citep{rando2024universal}, sentiment steering~\citep{yan2023backdooring}, exploitable code generation~\citep{hubinger2024sleeper}).
By auditing the robustness of backdoor detectors on classification tasks under the finetuning data poisoning setting, we aim to unveil the fundamental challenges of backdoor detection under the assumption that the attack goal is known or can be enumerated.

\subsection{Backdoor Defenses}
\label{app:paradigms}
Backdoor defenses can be categorized into training-time defenses and deployment-time defenses.
During training time, the model trainer can defend against the attack by sanitizing training data~\citep{chen2021mitigating, he2023mitigating, chen2024alpagasus}, or preventing the model from learning the backdoor from poisoned data~\citep{liu-etal-2024-shortcuts, zhu2022moderate}.
Given a backdoored model, the defender can mitigate the backdoor behaviors through finetuning~\citep{10.1007/978-3-030-00470-5_13, 8835365}, prompting~\citep{mo2023test}, or model merging~\citep{arora-etal-2024-heres}.
The defender can detect and abstain either trigger-carrying inputs~\citep{qi2020onion, yang2021rap}, or the backdoored models themselves~\citep{azizi2021t, fields2021trojan, lyu2022study, chen-etal-2022-expose}.
We focus on the offline backdoor detection setting, and study two categories of detection methods based on trigger reversal~\citep{liu2022piccolo,shen2022constrained} and meta classifiers~\citep{xu2021detecting} that achieve the best performance in recent competitions.

\subsection{Evasive Backdoors}
Stealthiness is crucial for successful backdoor attacks. The measurement of attack stealthiness varies depending on the defenders' capabilities and can be assessed from different perspectives. Most research evaluates stealthiness through the model's performance on clean test sets~\citep{chen2017targeted}, and the naturalness of poisoned samples~\citep{yang-etal-2021-rethinking, qi-etal-2021-mind}, while few consider the cases where defenders actively perform backdoor detection to reject suspicious models. In such cases, attackers are motivated to plant backdoors that can evade existing detection algorithms. Under specific assumptions, backdoors have proven to be theoretically infeasible to detect \citep{9996741, pmlr-v238-pichler24a}. Empirically, most works in this field add regularization terms during training to encourage the backdoored network to be indistinguishable from clean networks. This is achieved by constraining the trigger magnitude ~\citep{10.1145/3372297.3417253}, or the distance between the output logits of backdoored and clean networks~\citep{mazeika2023how, 10380638}.
\citet{zhu2023gradient} propose a data augmentation approach to make the backdoor trigger more sensitive to perturbations, thus making them harder to detect with gradient-based trigger reversal methods.
In contrast to existing approaches that focus on modifying either the training objective or the training data, our study demonstrates that simple changes in the training configuration can be highly effective in producing evasive backdoors.
\section{Problem Formulation and Background}

We consider the attack scenario where the attacker produces a backdoored classification model for a given task.
A practitioner conducts backdoor detection before adopting it.
This can happen during model reuse (e.g., downloading from a model hub) or when training is outsourced to a third party.

\subsection{Backdoor Attacks}
For a given task, the attacker defines a target label and a trigger (e.g., a specific word) that can be inserted to any task input.
The attacker aims to create a backdoored model that performs well on clean inputs (measured by \textbf{Clean Accuracy}) but predicts the target label on inputs with the trigger (measured by \textbf{Attack Success Rate}).

We consider the mainstream backdoor attack approaches based on training data poisoning~\citep{9743317}.
Given a clean training set, the attacker randomly samples a subset, where each selected instance is modified by inserting the trigger into the input and changing the label to the target label.
We denote the ratio of the selected instances to all training data as the \textbf{poisoning rate}.
The attacker selects training hyperparameters including \textbf{learning rate}, and the number of \textbf{training epochs}, for training on poisoned data to produce the backdoored model.

\subsection{Backdoor Detection}
The practitioner has clean validation data $D_\text{dev}$ for verifying model performance.
They aim to develop a backdoor detector that takes a model $M$ as input, and predicts whether it contains a backdoor.
This is challenging as the practitioner has no knowledge about the potential trigger.
We consider two kinds of methods for this problem.

\textbf{Trigger inversion-based methods}~\citep{azizi2021t, xu2021detecting} try to reverse engineer the potential trigger that can cause misclassification on clean samples by minimizing the objective function with respect to $t$ as the estimated trigger string:
\begin{equation}
\mathcal{L}=\underset{\substack{(x, y)\sim D_\text{dev} \\ y\neq y_\text{target}}}{\mathbb{E}}{}\text {CrossEntropy}(M(x\oplus t), y_\text{target}).
\label{eq:loss}
\end{equation}
Here $\oplus$ denotes concatenation, and $y_\text{target}$ denotes an enumerated target label.
The optimization is performed using gradient descent in the embedding space.
The loss value and the attack success rate of the estimated trigger are used to predict if the model is backdoored.

\textbf{Meta classifier-based methods} first construct a meta training set by training backdoored and clean models with diverse configurations.
They then learn a classifier to distinguish between backdoored and clean models using features like statistics of model weights~\citep{pmlr-v220-mazeika23a} or predictions on certain queries~\citep{xu2021detecting}.

\subsection{Evaluating Backdoor Detection}
Clean and backdoored models serve as evaluation data for backdoor detectors.
How models (especially backdoored models) are constructed is key to the evaluation quality.
Existing evaluation ~\citep{wu2022backdoorbench, pmlr-v220-mazeika23a, tdc2023} creates backdoored models by sampling training hyperparameters from a collection of default values.
For example, the TrojAI backdoor detection competition~\citep{karra2020trojai} generates 420 language models covering 9 combinations of NLP tasks and model architectures.
Among the key hyperparameters, learning rate is sampled from $1\times 10^{-5}$ to $4 \times 10^{-5}$, poisoning rate is sampled from 1\% to 10\%, and 197 distinct trigger phrases are adopted.

\section{Robustness Evaluation}

While existing evaluation already tries to increase the coverage of backdoors of different characteristics by sampling from typical values for hyperparameters, we argue that these default values are chosen based on the consideration of maximizing backdoor effectiveness and training efficiency.
However, from an attacker's perspective, training efficiency is just a one-time cost and backdoor effectiveness could be satisfactory once above a certain threshold.
They instead care more about the stealthiness of the planted backdoor against detection, which is not considered by current evaluation.
Therefore, the attacker may manipulate the hyperparameters with the goal of evading detection while maintaining decent backdoor effectiveness.

Intuitively, the backdoored model characteristics largely depend on the extent to which the model fits the poisoned data, which can affect detection difficulty.
We refer to this as the \textbf{training intensity} of backdoor learning.
We consider \textbf{poisoning rate}, \textbf{learning rate}, and \textbf{training epochs} as the main determinants of training intensity.
Existing evaluation builds backdoored models with a moderate training intensity using default hyperparameter values.
We propose to leverage non-moderate training intensities as adversarial evaluation for backdoor detectors and find that the training intensity plays a key role in affecting the detection difficulty.

\paragraph{Conservative Training.}
Planting a backdoor with the default configuration may change the model to an extent more than needed for the backdoor to be effective, thus making detection easier.
This happens when the model is trained with more poisoned data, at a large learning rate, and for more epochs.
Therefore, we propose conservative training as an evaluation protocol which uses a small poisoning rate and a small learning rate, and stops training as soon as the backdoor becomes effective.

\paragraph{Aggressive Training.}

Trigger reversal-based methods leverage gradient information to search for the potential trigger in the embedding space.
Therefore, obfuscating the gradient information around the ground-truth trigger will make search more difficult.
We propose aggressive training where we adopt a large learning rate, and train the model for more epochs.
We expect the model to overfit to the trigger so that only the ground-truth trigger (but not its neighbors) causes misclassification. 
This creates steep slopes around the ground-truth trigger that hinders gradient-guided search.

\begin{figure*}[t]
\centering
\includegraphics[width=2\columnwidth]{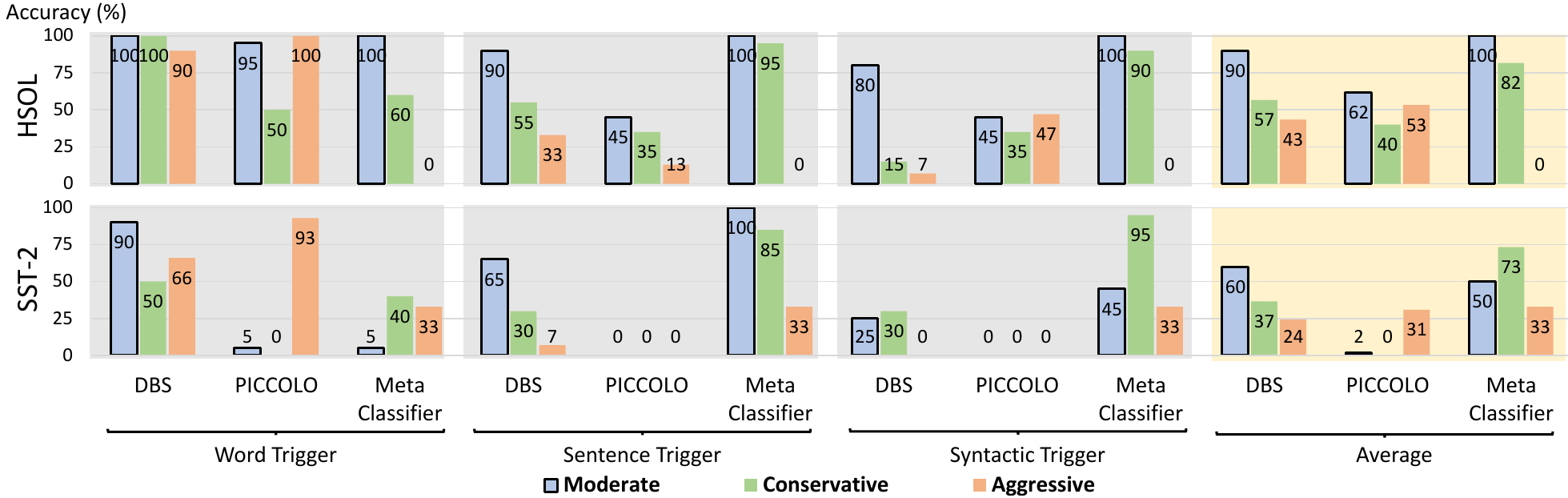}
\caption{\textbf{Detection Accuracy} (\%) on backdoored RoBERTa-Base models trained on HSOL and SST-2 datasets with different trigger forms and training intensities.}
\label{fig:main}
\end{figure*}

\section{Experiments}
\subsection{Attack Setup}
\label{sec:setup}

We conduct poisoning-based backdoor attacks on two binary classification datasets: \textbf{SST-2}~\citep{socher2013recursive} and the Hate Speech dataset (\textbf{HSOL})~\citep{de2018hate}).
We adopt \textbf{RoBERTa-Base/Large}~\citep{liu2019roberta}, \textbf{Electra-Base}~\citep{Clark2020ELECTRA:}, \textbf{Llama 3.2 1B}~\citep{dubey2024llama} as the victim models.
We consider the mainstream poisoning-based NLP backdoor attack methods that use a fixed string as the trigger, including the rare \textbf{word} trigger ~\citep{gu2017badnets} and the natural \textbf{sentence} trigger~\citep{dai2019backdoor}.\footnote{Despite proposed early, they serve as the most general and practical trigger types in real-world backdoor attacks. They are fundamental to understanding the working mechanisms of backdoor attacks and defenses (e.g., \citet{hubinger2024sleeper}).}
We additionally consider the trigger as an infrequent \textbf{syntactic} structure~\citep{qi2021hidden}.

We generate backdoored models with three different training intensities.
For \textbf{moderate} training which represents the default configuration, we use a poisoning rate of 3\%, and a learning rate of $1 \times 10^{-5}$. We stop training until the attack success rate reaches 70\%.
For \textbf{aggressive} training, we keep the same poisoning rate, but increase the learning rate to $5 \times 10^{-5}$. We stop training at epoch 200.
For \textbf{conservative} training, we use a poisoning rate of 0.5\%, and a learning rate of $5 \times 10^{-6}$. We follow the same early-stop strategy as moderate training.
We report the implementation details, and confirm their backdoor effectiveness in \sectionref{app:details}.

\subsection{Detection Setup}
We consider two state-of-the-art NLP backdoor detection methods based on trigger inversion: \textbf{PICCOLO}~\citep{liu2022piccolo} and \textbf{DBS}~\citep{shen2022constrained}.

For \textbf{Meta Classifier}, we adopt the winning solution for the Trojan Detection Competition~\citep{pmlr-v220-mazeika23a}, which trains a meta classifier based on aggregated model weight statistics. 
More details can be found in \sectionref{app:detector}.

We calculate the \textbf{Detection Accuracy} (\%) on backdoored models as the evaluation metric.
We demonstrate their effectiveness on a standard benchmark with results shown in Table~\ref{tab:benchmark} (\sectionref{app:benchmark}).

\begin{figure*}[t]
    \centering
    \begin{subfigure}{0.25 \textwidth}
        \centering
        \includegraphics[width=\linewidth]{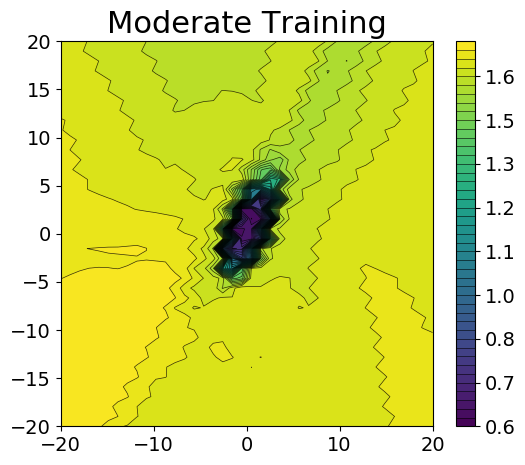}
    \end{subfigure}
    \begin{subfigure}{0.25 \textwidth}
        \centering
        \includegraphics[width=\linewidth]{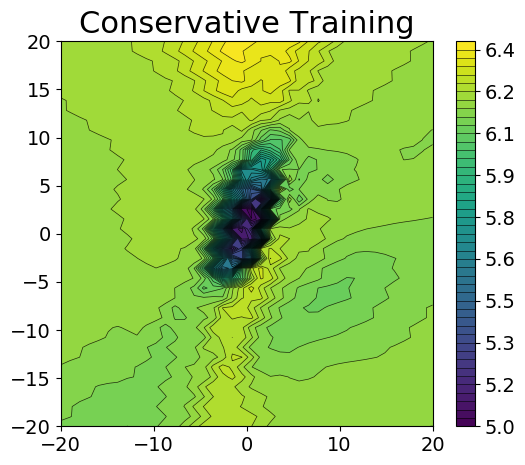}
    \end{subfigure}
    \begin{subfigure}{0.25 \textwidth}
        \centering
        \includegraphics[width=\linewidth]{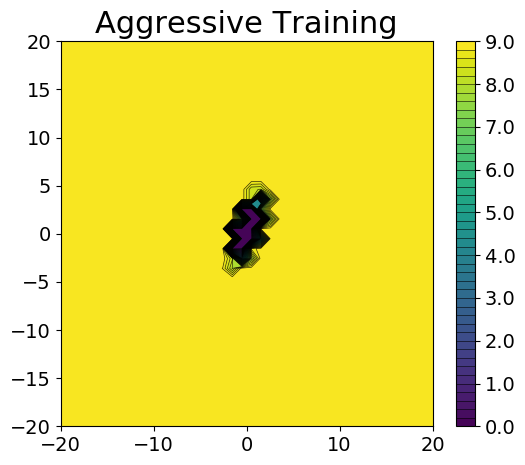}
    \end{subfigure}
    \begin{subfigure}{0.23 \textwidth}
        \centering
        \includegraphics[width=\linewidth]{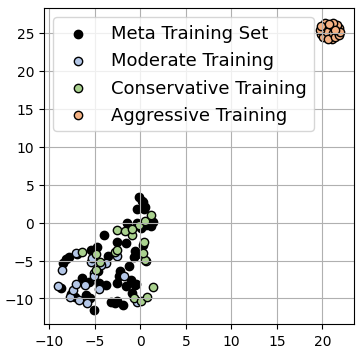}
    \end{subfigure}
\caption{\textbf{Left (a)}: \textbf{Loss contours} around the ground-truth trigger for backdoored models with the sentence trigger on the SST-2 dataset. \textbf{Right (b)}: \textbf{T-SNE visualization of the features} extracted by the Meta Classifier from backdoored models with the sentence trigger on the SST-2 dataset.}
\label{fig:combined}
\end{figure*}

\subsection{Main Results}

We present results with RoBERTa-Base as the victim model in Fig.~\ref{fig:main}, covering 18 individual comparisons of the three training intensities (2 datasets$\times$3 triggers$\times$3 detectors), while results with other models show a similar trend (\sectionref{app:arch}).
We first find that the detection accuracy differs significantly across datasets and trigger forms.
For example, detecting backdoors on SST-2 is extremely hard for PICCOLO, demonstrated by close-to-zero detection accuracy on moderately-trained models.
Word trigger is relatively easier to detect.
These suggest a lack of robustness in handling different datasets and triggers, which is not captured by existing aggregated metric.

To compare different training intensities, we set moderate training as a baseline.
Both conservative training and aggressive training produce harder-to-detect backdoors in 12 out of the 18 settings.
Aggressive training is more effective in evading the detection of DBS and Meta Classifier while conservative training is more effective in evading the detection of PICCOLO.
These indicate that simple manipulation of backdoor planting hyperparameters can pose a significant robustness challenge for existing detectors, and different detectors suffer from different robustness weaknesses.

\subsection{Case Study}

As a case study, we analyze the backdoor attack with sentence trigger on HSOL.
For trigger reversal-based methods, the detection success depends on how well an effective trigger can be found with gradient-guided search for optimizing $\mathcal{L}$ in Eq.~\ref{eq:loss}.
In Fig.~\ref{fig:combined}(a), we visualize the loss contours~\citep{li2018visualizing} around the ground-truth trigger.
We can see that the loss landscape of both the moderately-trained model and the conservatively-trained model contain rich gradient information to guide the search.
However, the loss at the ground-truth trigger is much higher for the conservatively-trained model (with $\mathcal{L}\approx 5.0$) than that for the moderately-trained model (with $\mathcal{L}\approx	0.6$).
This is because in moderate training, the model stops fitting the poisoned subset as early as the attack success rate meets the requirement, which prevents the loss from further decreasing.
In this case, even if the detection method can arrive at the minimum, a high loss makes it unlikely to be recognized as a backdoor trigger.
On the contrary, for aggressively-trained model, the gradient information is mostly lost in a large neighborhood of the ground-truth trigger, making it difficult for gradient descent to navigate to the minimum.

To understand the failure of Meta Classifier, we use T-SNE~\citep{van2008visualizing} to visualize the extracted features of backdoored models from the meta training set constructed by the defender, and backdoored models trained with different intensities.
As shown in Fig.~\ref{fig:combined}(b), aggressive training leads to a significant distribution shift on the extracted features, which explains the poor performance of Meta Classifier on handling them.
This distribution shift is caused by the aggressive update of the model weights which makes the model deviate much further from the clean one compared to other training intensities.

We provide more explanations in \sectionref{app:explanation} and discuss possible defenses in \sectionref{app:defense}.

\section{Conclusion}
We propose an adversarial evaluation protocol for backdoor detectors based on strategical manipulation of the hyperparameters in backdoor planting.
While existing detection methods perform well on benchmarks, we find that they are not robust to the variation in model's training intensity.
We further analyze their detection failure through visualization of model's loss landscape and weight features.
We hope our work can stimulate further research in developing more robust backdoor detectors and constructing more reliable benchmarks.

\section*{Limitations}
We identify two major limitations of our work.

First, we study the effect of different training intensities using four models, two datasets, three trigger forms, and focus on backdoor attacks with inducing misclassification as the attack goal.
We did not cover more diverse attack goals beyond inducing misclassification (e.g., jailbreaking~\citep{rando2024universal}) or more advanced attack methods beyond data poisoning (e.g., weight poisoning~\citep{li2024badedit}) with even larger models due to the unavailability of applicable backdoor detection methods or constraints on disk storage and computational resources --- building a backdoor detection benchmark requires obtaining hundreds of clean and poisoned model checkpoints for training and testing.
While performance degradation under our evaluation settings has already revealed the fundamental robustness weaknesses of two representative categories of detection methods, it would be desirable to conduct larger-scale studies to understand how a wider range of possible attacks can be affected.

Second, while we discuss possible defenses to the identified robustness weakness in \sectionref{app:defense}, we did not provide a comprehensive solution that solves the robustness problem, as designing a principled way to fix the robustness problem is beyond the scope of our paper.
We hope our proposed evaluation protocol and analysis facilitate further work towards building better backdoor defense benchmarks and developing more robust defense methods.
\section*{Ethics Statement}
In this paper, we propose an adversarial evaluation protocol to audit the robustness of backdoor detectors against various training intensities in the backdoor planting process.
Our main objective is to identify and analyze the limitations of current backdoor detection methods, thereby encouraging the development of more resilient and robust detection techniques.

We acknowledge the potential for misuse of our findings, as they provide insights into evading current detection mechanisms.
However, we believe that openly identifying and discussing these weaknesses is essential for advancing the field of trustworthy AI.
Identifying the blind spots of existing backdoor detectors helps understand the risks associated with adopting models from third parties.
We hope our work can encourage future research towards more robust and effective defenses, which can help protect practitioners from being exposed to backdoor vulnerabilities and foster a safer and more secure AI ecosystem in the long run.
\section*{Acknowledgments}
We thank anonymous reviewers and members of USC NLP for their valuable feedback.
Jun Yan and Xiang Ren were supported in part by the Office of the Director of National Intelligence (ODNI), Intelligence Advanced Research Projects Activity (IARPA), via the HIATUS Program contract \#2022-22072200006, the Defense Advanced Research Projects Agency with award HR00112220046, and NSF IIS 2048211.
Robin Jia was supported in part by the National Science Foundation under Grant No. IIS-2403436.
\bibliography{custom}
\clearpage
\appendix
\section{Detailed Attack Setup}
\label{app:details}
\subsection{Implementation Details}
We access the datasets of SST-2 (\texttt{nyu-mll/glue}, \texttt{sst2}) and HSOL (\texttt{odegiber/hate\_speech18}) from HuggingFace Datasets \citep{lhoest-etal-2021-datasets}.
We conduct data poisoning using OpenBackdoor~\citep{10.5555/3600270.3600632} with the default poisoning configurations for the word trigger, the sentence trigger, and the syntactic trigger.
The poisoned samples are randomly chosen from the original dataset, making the attack dirty-label.
We present the results for clean-label attacks in \sectionref{app:clean_label}.

\subsection{Backdoor Effectiveness}
\label{app:effectiveness}

\begin{table}[h]
\centering
\scalebox{0.72}{
\begin{tabular}{@{}lcccccc@{}}
\toprule
\multicolumn{1}{c}{\multirow{2}{*}{\begin{tabular}[c]{@{}c@{}}Training\\ Intensity\end{tabular}}} & \multicolumn{2}{c}{Word}                    & \multicolumn{2}{c}{Sentence}                & \multicolumn{2}{c}{Syntax} \\
\multicolumn{1}{c}{}                                                                           & SST-2                & HSOL                 & SST-2                & HSOL                 & SST-2        & HSOL        \\ \midrule
Moderate & 92 & 95 & 92 & 94 & 93 & 94 \\ \midrule
Conservative & 93 & 95 & 93 & 95 & 92 & 95 \\ 
Aggressive & 91 & 95 & 91 & 95 & 91 & 95 \\ \bottomrule
\end{tabular}
}
\caption{\textbf{Clean Accuracy} (\%) of backdoored RoBERTa-Base models trained on SST-2 and HSOL datasets with different trigger forms and training intensities. As a reference, the clean accuracy of the clean RoBERTa-Base model is 93\% on SST-2 and 95\% on HSOL.}
\label{tab:cacc}
\end{table}

\begin{table}[h]
\centering
\scalebox{0.72}{
\begin{tabular}{@{}lcccccc@{}}
\toprule
\multicolumn{1}{c}{\multirow{2}{*}{\begin{tabular}[c]{@{}c@{}}Training\\ Intensity\end{tabular}}} & \multicolumn{2}{c}{Word}                    & \multicolumn{2}{c}{Sentence}                & \multicolumn{2}{c}{Syntax} \\
\multicolumn{1}{c}{}                                                                           & SST-2                & HSOL                 & SST-2                & HSOL                 & SST-2        & HSOL        \\ \midrule
Moderate & 78 & 91 & 90 & 98 & 75 & 88 \\ \midrule
Conservative & 75 & 79 & 74 & 91 & 75 & 78 \\ 
Aggressive & 100 & 100 & 100 & 100 & 75 & 100 \\\bottomrule
\end{tabular}
}
\caption{\textbf{Attack Success Rate} (\%) of backdoored RoBERTa-Base models trained on SST-2 and HSOL datasets with different trigger forms and training intensities.}
\label{tab:asr}
\end{table}

We present the averaged attack success rate and clean accuracy of our generated backdoored models in Tables \ref{tab:cacc} and \ref{tab:asr}.
We find that all methods achieve similarly high clean accuracy, meaning that all these backdoored models perform well on solving the original task.
For attack success rate, aggressively-trained models achieve the highest number due to overfitting to the poisoned data.
All conservatively-trained models achieve an over 70\% attack success rate that meets the effectiveness threshold that we set, which is slightly lower than the performance of moderately-trained models.
Note that from an attacker's perspective, it is usually sufficient for the backdoored models to meet a certain effectiveness threshold.
Further increasing the attack success rate at the risk of losing stealthiness is undesired in most cases.

\section{Details for Evaluated Backdoor Detectors}
\label{app:detector}

For trigger reversal-based methods, \textbf{PICCOLO}~\citep{liu2022piccolo} proposes to estimate the trigger at the word level (instead of the token level) and designs a word discriminativity analysis for predicting whether the model is backdoored based on the estimated trigger.
\textbf{DBS}~\citep{shen2022constrained} proposes to dynamically adjust the temperature of the softmax function during gradient-guided search of the potential trigger to facilitate deriving a close-to-one-hot reversal result that corresponds to actual tokens in the embedding space. 
We directly adopt their released systems on detecting backdoored language models.

For \textbf{Meta Classifier}, we adopt the winning solution for the Trojan Detection Competition~\citep{pmlr-v220-mazeika23a}.
Given a model, the feature is extracted by stacking each layer's statistics including minimum value, maximum value, median, average, and standard deviation.
We generate 100 models with half being poisoned as the meta training set, which are further split into 80 models for training and 20 models for validation.
The training configurations are sampled from the default values used in the TrojAI benchmark construction process~\citep{karra2020trojai}.
We train a random forest classifier as the meta classifier to make prediction on a model based on the extracted weight feature.
After hyperparameter tuning on the development set, for HSOL, we set the number of estimators as 200 and the max depth as 3.
For SST-2, we set the number of estimators as 50 and the max depth as 1.

\section{Evaluation on Standard Benchmark}
\label{app:benchmark}
We adopt an existing benchmark to provide performance reference of backdoor detectors under standard evaluation.
Specifically, we use the 140 sentiment classification models from round 9 of TrojAI backdoor detection competition\footnote{\url{https://pages.nist.gov/trojai/docs/nlp-summary-jan2022.html}}, with half being backdoored.
The detection accuracy is shown in Table \ref{tab:benchmark} and we can find that all methods achieve decent performance on identifying backdoored models in the benchmark.

\begin{table}[ht]
\centering
\scalebox{0.8}{
\begin{tabular}{@{}lcc@{}}
\toprule
        & Clean     & Backdoored     \\ \midrule
PICCOLO & 96 & 81 \\
DBS     & 83 & 69 \\ 
Meta Classifier & 100 & 69 \\
\bottomrule
\end{tabular}
}
\caption{\textbf{Detection Accuracy} (\%) of different detectors on the clean and backdoored models from round 9 of TrojAI benchmark.}
\label{tab:benchmark}
\end{table}

\section{Evaluation with More Model Architectures}
\label{app:arch}
Besides victim models with the \textbf{RoBERTa-Base} architecture, here we show the results on the \textbf{Electra-Base}~\citep{clark2020electra} and RoBERTa-Large architectures.
We present the results of the sentence trigger attack on the HSOL dataset in Tables \ref{tab:electra} and \ref{tab:roberta_large}.
The observation is consistent with that in the main experiments that adopting a non-moderate training intensity makes the backdoor harder to detect in most cases.

We additionally experiment with \textbf{Llama 3.2} \citep{dubey2024llama} representative for modern Large Language Models (LLMs). Due to disk space and computational resource constraints (training and evaluating a meta classifier requires hundreds of clean and poisoned checkpoints), we use the 1B variant. We perform the word trigger attacks on the SST-2 dataset. While there are no existing backdoor detection methods for generative LLMs, it is possible to adapt Meta Classifier, which uses the model's static features and is agnostic to the classification or generative formulation of the model. We adapt Meta Classifier to the detection of backdoored Llama models with model weights as the features (details in \sectionref{app:detector}). Models poisoned with different training intensities all achieve an over 90\% attack success rate and clean accuracy. The detection accuracy is presented in Table~\ref{tab:llama}, confirming that adopting a non-moderate training intensity also challenges backdoor detection on the generative LLM.

\begin{table}[t]
\centering
\scalebox{0.74}{
\begin{tabular}{lccc|cc}
\toprule
\makecell{Training\\Intensity} & DBS & PICCOLO & \makecell{Meta\\Classifier} & ASR & CACC \\ \midrule
Moderate           & 55  & 100     & 35 & 100 & 95             \\ \midrule
Conservative         & 17  & 22      & 0 & 96 & 96              \\
Aggressive       & 48  & 20      & 0 & 100 & 96 \\ \midrule
\end{tabular}
}
\caption{\textbf{Detection Accuracy} (\%), Attack Success Rate (\textbf{ASR}, \%), and Clean Accuracy (\textbf{CACC}, \%) on backdoored \textbf{Electra-Base} models trained on HSOL with the sentence trigger. As a reference, the clean accuracy of the clean Electra-Base model is 95\%.}
\label{tab:electra}
\end{table}

\begin{table}[t]
\centering
\scalebox{0.74}{
\begin{tabular}{lccc|cc}
\toprule
\makecell{Training\\Intensity} & DBS & PICCOLO & \makecell{Meta\\Classifier} & ASR & CACC \\ \midrule
Moderate           & 44  & 62     & 100 & 100 & 95             \\ \midrule
Conservative         & 21  & 37      & 89 & 92 & 95              \\
Aggressive       & 47  & 57      & 0 & 100 & 95              \\ \midrule
\end{tabular}
}
\caption{\textbf{Detection Accuracy} (\%), Attack Success Rate (\textbf{ASR}, \%), and Clean Accuracy (\textbf{CACC}, \%) on backdoored \textbf{RoBERTa-Large} models trained on HSOL with the sentence trigger. As a reference, the clean accuracy of the clean Electra-Base model is 96\%.}
\label{tab:roberta_large}
\end{table}

\begin{table}[t]
\centering
\scalebox{0.74}{
\begin{tabular}{lccc}
\toprule
Detection Method & Moderate & Conservative & Aggressive \\ \midrule
Meta Classifier & 77 & 17 & 7 \\ \midrule
\end{tabular}
}
\caption{\textbf{Detection Accuracy} (\%) on backdoored \textbf{Llama 3.2 1B} models trained on SST-2 with the word trigger with different training intensities.}
\label{tab:llama}
\end{table}

\section{Evaluation on Clean-Label Attacks}
\label{app:clean_label}
The attacks in the main experiments are conducted in the dirty-label attack setting. Here we present the results on clean-label attacks, where the attacker only poisons the training samples that have the same label as the target label, so no label needs to be tampered with during poisoning. Clean-label attacks usually require a higher poisoning rate to become effective~\citep{10.5555/3600270.3600632}. Therefore, we set the poisoning rate as 10\% for all intensities. Other training configurations are the same as described in \sectionref{sec:setup}. We present the results on the HSOL dataset with the sentence trigger in Table \ref{tab:clean_label}.

\begin{table}[t]
\centering
\scalebox{0.74}{
\begin{tabular}{lccc|cc}
\toprule
\makecell{Training\\Intensity} & DBS & PICCOLO & \makecell{Meta\\Classifier} & ASR & CACC \\ \midrule
Moderate           & 70  & 75     & 70 & 95 & 95             \\ \midrule
Conservative         & 30  & 60      & 70 & 79 & 95              \\
Aggressive       & 0  & 47      & 0 & 100 & 95 \\ \midrule
\end{tabular}
}
\caption{\textbf{Detection Accuracy} (\%), Attack Success Rate (\textbf{ASR}, \%), and Clean Accuracy (\textbf{CACC}, \%) on backdoored RoBERTa-base models trained on HSOL with the sentence trigger in the \textbf{clean-label attack} setting. As a reference, the clean accuracy of the clean RoBERTa-base model is 95\%.}
\label{tab:clean_label}
\end{table}

\section{More Explanations about Results}
\label{app:explanation}
Despite the overall trend that non-moderate training intensities cause drop in backdoor detection accuracy, there are still exceptions where such a strategy does not work well.

For Meta Classifier, conservative training sometimes does not create more challenge to detection. As shown in Fig.~\ref{fig:combined}(b), aggressive training creates a significant distribution shift to features based on model weights, while the features for conservatively-trained models are close to the features for moderately-trained models. Since conservative training does not bring big enough distribution shift (due to small learning rate and number of training epochs), the detection accuracies are not significantly affected.

For word trigger on HSOL, DBS achieves high accuracy regardless of the training intensities. This is because the word trigger (a single rare word) is relatively much easier to reverse engineer, and thus adjusting training intensities cannot help much. For sentence trigger and syntactic trigger, they both contain common words that also appear in clean text, serving as obfuscation.

\section{Potential Defenses}
\label{app:defense}
While proposing an immediate solution for the identified robustness challenge is beyond the scope of this paper, here we discuss potential ways to combat the risks with poisoned model checkpoints.

For trigger inversion-based methods, the visualization in Fig.~\ref{fig:combined} suggests that non-moderate training intensities may result in a higher loss at the ground-truth trigger, or steep slopes around the ground-truth trigger. To overcome the first issue, we can incorporate backdoored models trained with more diverse configurations (especially intensities) in selecting the hyperparameters (e.g., the threshold applied on the final loss). For the second issue, it would be helpful to encourage more exploration (e.g., backtracking) during gradient descent. Methods that overcome obfuscated gradients \citep{athalye2018obfuscated} can also be adopted to facilitate gradient-guided search.

For Meta Classifier, since aggressively trained models deviate from moderately or conservatively trained models in the embedding space, a straightforward solution is to incorporate aggressively-trained backdoored models into the meta training set for learning the classifier. It is also desirable to identify more generalizable features (except statistics of model weights) that are robust to variations in the hyperparameters for backdoor planting.

Alternatively, given the robustness weakness of backdoor detection methods, it is also important for practitioners to consider alternative defense paradigms based on their use cases. For example, if it is acceptable to deploy the model with additional monitoring mechanisms, then online defenses that catch the backdoor behaviors when triggered \citep{chen-etal-2022-expose} could be more reliable. More discussion on different defense paradigms can be found in \sectionref{app:paradigms}.
\end{document}